\documentclass[runningheads]{llncs}
\usepackage{hyperref}
\usepackage{float}
\usepackage{cite}
\usepackage{graphics,graphicx}
\usepackage{xcolor}
\usepackage{soul}
\usepackage{amsmath,mathtools}
\usepackage[utf8]{inputenc}
\usepackage[acronym,smallcaps]{glossaries}
\usepackage{subfigure}
\usepackage{booktabs}

\newacronym{rl}{RL}{\textit{Reinforcement Learning}}
\newacronym{mfrl}{MFRL}{\textit{Model-Free Reinforcement Learning}}
\newacronym{ml}{ML}{\textit{Machine Learning}}
\newacronym{marl}{MARL}{\textit{Multi-agent Reinforcement Learning}} 
\newacronym{il}{IL}{\textit{Independent Learners}} 
\newacronym{jal}{JAL}{\textit{Joint-Action Learners}} 

\newacronym{vsss}{VSSS}{\textit{IEEE Very Small Size Soccer}}
\newacronym{ssl}{SSL}{\textit{Small Size League}}
\newacronym{fira}{FIRA}{\textit{Federation of International Robot-Sport Association}}
\newacronym{mdp}{MDP}{\textit{Markov Decision Process}}
\newacronym{larc}{LARC}{\textit{Latin American Robotics Competition}}

\newacronym{dqn}{DQN}{\textit{Deep Q Network}}
\newacronym{ddpg}{DDPG}{\textit{Deep Deterministic Policy Gradient}}
\newacronym{sac}{SAC}{\textit{Soft Actor Critic}}
\newacronym{drl}{DRL}{\textit{Deep Reinforcement Learning}}
\newacronym{fnn}{FNN}{\textit{Fuzzy Neural Network}}
\newacronym{ac}{AC}{\textit{Actor Critic}}
\newacronym{ir}{IR}{\textit{Infrared sensor}}

\newacronym{model}{VSSS-RL}{\textit{Very Small Size Soccer Reinforcement Learning}}

\newcommand{\rsim}{rSim}

\newacronym{ode}{ODE}{\textit{Open Dynamics Engine}}
\newacronym{gym}{gym}{\textit{OpenAi Gym Toolkit}}
\newacronym{ou}{OU}{Ornstein-Uhlenbeck process}
\newacronym{si}{SI}{International System of Units}

\title{
rSoccer: A Framework for Studying Reinforcement Learning in Small and Very Small Size Robot Soccer
}
\titlerunning{rSoccer Framework}
\author{Felipe B. Martins \and
Mateus G. Machado \and
Hansenclever F. Bassani \and
Pedro H. M. Braga \and
Edna S. Barros
}
\authorrunning{F. Martins et al.}
\institute{Centro de Informática - Universidade Federal de Pernambuco, Av. Jornalista Anibal Fernandes, s/n - CDU 50.740-560, Recife, PE, Brazil.\\
\email{\{fbm2, mgm4, hfb, phmb4, ensb\}@cin.ufpe.br}\\
}
\begin{document}

\maketitle

\thispagestyle{plain}
\thispagestyle{empty}

\begin{abstract}
Reinforcement learning is an active research area with a vast number of applications in robotics, and the RoboCup competition is an interesting environment for studying and evaluating reinforcement learning methods. A known difficulty in applying reinforcement learning to robotics is the high number of experience samples required, being the use of simulated environments for training the agents followed by transfer learning to real-world (sim-to-real) a viable path. This article introduces an open-source simulator for the IEEE Very Small Size Soccer and the Small Size League optimized for reinforcement learning experiments. We also propose a framework for creating OpenAI Gym environments with a set of benchmarks tasks for evaluating single-agent and multi-agent robot soccer skills. We then demonstrate the learning capabilities of two state-of-the-art reinforcement learning methods as well as their limitations in certain scenarios introduced in this framework. We believe this will make it easier for more teams to compete in these categories using end-to-end reinforcement learning approaches and further develop this research area.
\end{abstract}



\section{Introduction}
\label{sec:intro}

\gls{rl} \cite{sutton2018reinforcement}, in conjunction with the machine learning field, has obtained interesting results in progressively more complex decision-making competitive scenarios, such as learning how to play Atari games \cite{volodymyr2013playing}, Go \cite{silver2017mastering}, and Starcraft 2 \cite{vinyals2019alphastar}, achieving or even surpassing human-level performance. In robotics, \gls{rl} showed promising results in simulated and real-world environments, including approaches for motion planning, optimization, grasping, manipulation, and control \cite{christiano2016transfer, andrychowicz2018learning, levine2016end}.

Robot soccer competitions are an exciting field for researching and validating \gls{rl} usage, as it involves robotic systems capable of tackling challenging sequential decision-making problems in a cooperative and competitive scenario \cite{kim2004soccer}. Developing those systems can be very hard using traditional methods in which hard-coded behaviors need to foresee a multitude of possibilities in an unpredictable game such as soccer. 

\begin{figure}[!ht]
\begin{center}
    \subfigure[SSL]{\includegraphics[trim=0 40 0 88, clip, width=0.45\textwidth]{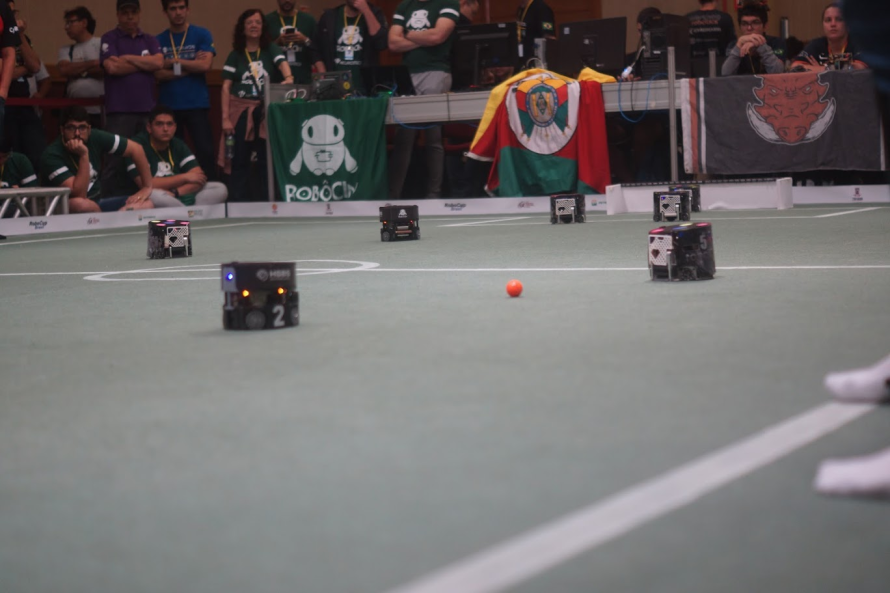}  \label{fig:ssl_real}}
    \subfigure[IEEE VSSS]{\includegraphics[trim=0 0 150 150, clip, width=0.45\textwidth]{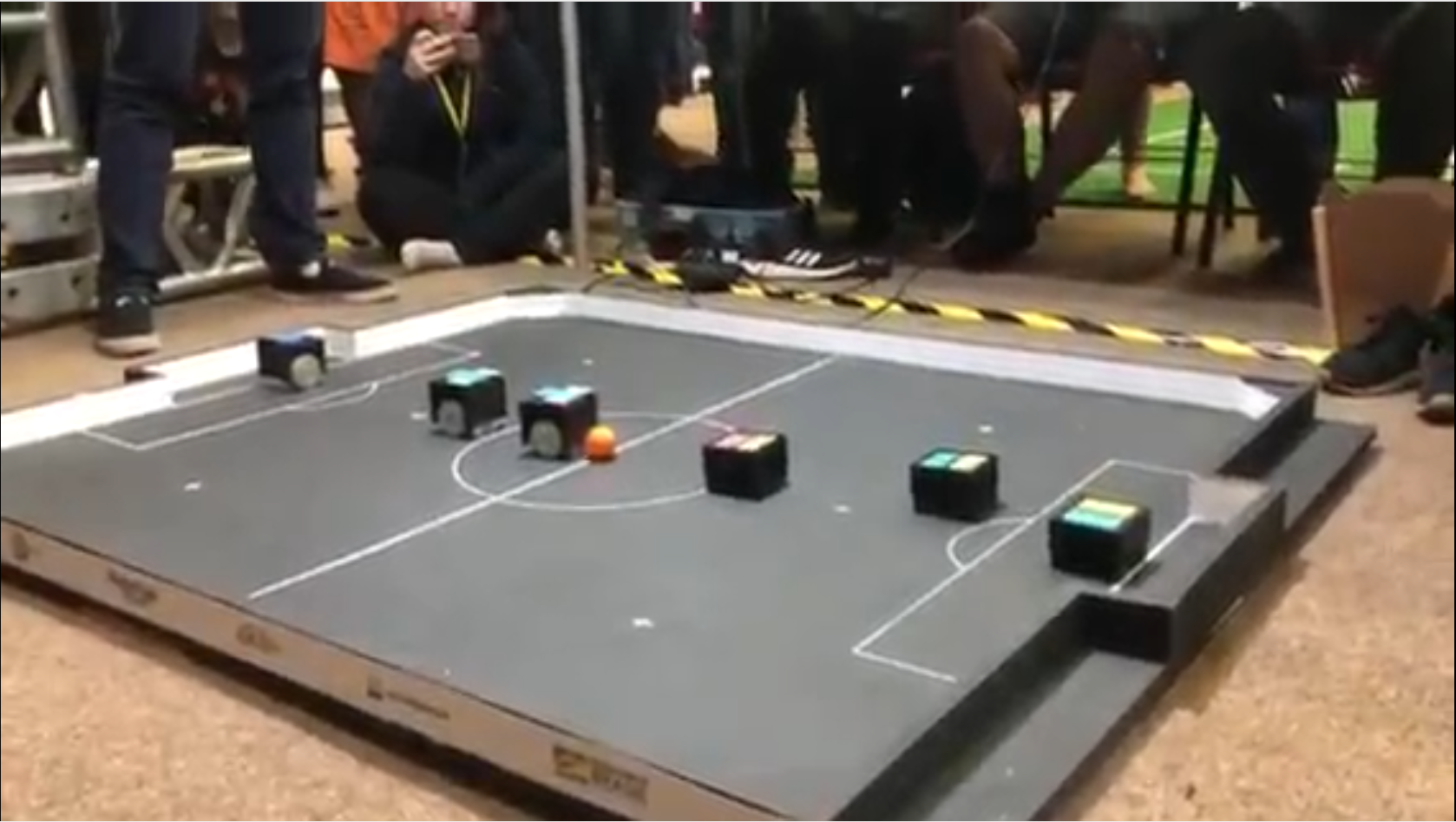} \label{fig:vss_real}}
\end{center}
    \caption{Competition environments of SSL and IEEE VSS.}
    \label{fig:real_envs}
\end{figure}

In the RoboCup \gls{ssl} competition, Fig. \ref{fig:ssl_real}, teams with up to eleven omnidirectional mobile robots compete against each other to score goals within a set of complex rules, such as limiting how far a robot can move with the ball, therefore requiring explicit cooperation. Previous works in this setting have successfully learned specific skills, such as moving to the ball, kicking, and defending penalties by using \gls{rl} control approaches \cite{yoon2015developing, zhu2019sslrl, zolanvari2019q}. However, those works did not make the learning environment available, which can hinder reproducibility. 

On the other hand, achieving an end-to-end control policy capable of cooperating on a robot soccer match is still an open problem, requiring even further research and development as the league evolves. Similarly, training a single policy capable of controlling a complete \gls{ssl} team is also challenging, as the total number of control actions increases with the number of robots.

The \gls{vsss} competition, Fig. \ref{fig:vss_real}, compared to the \gls{ssl}, establishes teams with three robots each, with a smaller field and robot sizes. The robot hardware does not have ball dribbling and kicking capabilities, and a match does not require explicit cooperation, by the rules. Although the differential drive robots used pose a more challenging path planning, the league can be seen as a simplified version of the \gls{ssl}. In this domain, earlier work applied \gls{rl} for learning specific skills \cite{duan2007application}. And more recently, Bassani et al. \cite{bassani2020framework} achieved 4th place in an international \gls{vsss} competition, using an end-to-end learned control without explicit cooperation and made the learning environments publicly available. Still, they do not support the \gls{ssl} setting and are not easily adaptable for different scenarios.

We consider that by supplying a simpler path towards creating and using RL SSL environments, the results achieved by Bassani et al. \cite{bassani2020framework} can be replicated on the SSL competition and be further used to encourage RL approaches in the SSL context. In summary, the contributions of the present work are the following:

\begin{enumerate}
    \item An open-source framework following the OpenAI Gym \cite{gym} standards for developing robot soccer \gls{rl} environments, modeling multi-agent tasks in competitive and cooperative scenarios;
    \item An open-source \gls{ssl} and \gls{vsss} robot soccer simulator, adapted from the grSim Simulator \cite{grsim}, focused on \gls{rl} use;
    \item A set of eight benchmark learning environments with a focus on reproducibility, for evaluating \gls{rl} algorithms in robot soccer tasks, including four tasks based on the RoboCup \gls{ssl} 2021 hardware challenges.
\end{enumerate}

The rest of this article is organized as follows: Section 2 presents related work on robot soccer simulators and environments. Section 3 describes the proposed framework. Section 4 introduces a set of benchmark environments created using the proposed framework. Section 5 presents the results, and finally, Section 6 draws the conclusions and suggests future work.

\section{Related Works}
\label{sec:related}
There is a large variety of \gls{rl} environments and frameworks on the literature which aim at allowing the easy reproduction of state-of-the-art \gls{rl} algorithms results, such as the OpenAI Gym \cite{gym}. However, the existing robot soccer environments lack the needed characteristics, such as extensibility to different scenarios, proper real-world robot simulation, and hard to reproduce results. Therefore, they do not apply to the RoboCup categories. These issues are discussed as follows.

\textbf{Suitable frameworks.} There are frameworks for simulating soccer matches such as the RoboCup's Soccer 2D \cite{robocup} and The Google Research Football Environment \cite{kurach2019google}. However, the actions defined are too high level. The DeepMind MuJoCo Multi-Agent Soccer Environment \cite{dmc_soccer} define low-level actions such as accelerating and rotating the body, but it is not related to a real robot soccer league. Bassani, \textit{et. al} \cite{bassani2020framework} proposed an framework for the \gls{vsss} setting, but it does not enable the creation of new scenarios. Although Robocup's Soccer 3D provides a low-level action and believable environment, there is no framework that enables the creation of scenarios.

\textbf{Simulator's purpose.} There are well known simulators for robot soccer competitions such as \gls{ssl} \cite{grsim} and \gls{vsss} \cite{firasim,vss_sdk}. They provide a real-time simulated environment with a rich graphical interface for developing robot soccer algorithms. However, the preferences for \gls{rl} are simulation speed and synchronous communication.

\textbf{Reproducibility issues.} Previous work achieved interesting results using \gls{rl} in robot soccer competition settings \cite{duan2007application, yoon2015developing, shi2018adaptive}. But they do not describe the environments and simulators used nor made them openly available, coupled with a lack of clearly defined tasks and availability of stable baseline implementations of robot soccer agents, poses several issues to the advancement of research in this field.

\section{rSoccer Gym Framework}

The proposed framework\footnote{Code available at \url{https://github.com/robocin/rSoccer}} is a tool for creating robot soccer environments ranging from simple single-agent tasks to complex multi-robot competitive cooperative scenarios.

\begin{figure}
  \includegraphics[width=0.95\linewidth]{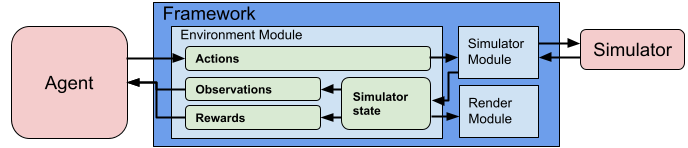}
  \caption{Framework modules architecture}
  \label{fig:architeture}
\end{figure}

It is defined by three modules: \textbf{simulator}, \textbf{environment}, and \textbf{render}. The simulator module describes the physics simulation. The environment module is designed to receive the agent action, communicate with the other modules, and return the new observations and rewards. The render module does the environment visualization. Fig. \ref{fig:architeture} illustrates the modules architecture. A set of data structures labeled \textbf{entities} are defined to enable a common communication between modules for every environment. The following subsections describe these modules and \textit{entities}. 

\subsection{Entities} \label{subsec_entities}
The \textit{entities} structures are standardized for consistency by defining positional values using the field center as a reference point. The units conform to the \gls{si}, except for robot angular position and speed values which are in degrees. The following \textit{entities} are defined:
\begin{itemize}
    \item \textit{Ball}: Contains the ball position and velocity values and is used both to read the current state or to set the initial position;
    \item \textit{Robot}: Contains a robot identification, flags, position, velocity, and wheel desired speed values. Used to read the current state, to set the initial position, or to send control commands;
    \item \textit{Frame}: Contains a \textit{Ball entity} and \textit{Robot entities} for each robot in the environment, structured in a way that each robot is easily indexable by team color and id. Used to store the complete state of the simulation;
    \item \textit{Field}: Contains specifications of the simulation values, such as the field and robot geometry and parameters.
\end{itemize}

\subsection{Simulator Module}
The simulator module carries out the environment physics calculations. It communicates directly with the environment module, receiving actions and returning the simulation state.

For physics calculations, we developed the \rsim{}\footnote{Code available at \url{https://github.com/robocin/rSim}} simulator specially for \gls{rl}. It was based on the grSim simulator \cite{grsim}, due to its reliable physical simulation, with the following modifications:

\begin{itemize}
    \item Removal of graphical interfaces to increase performance, reduce memory usage, and ease server deployment on headless servers;
    \item Synchronous operation for more consistent training results as in an asynchronous setting the synchronization between agents and simulator may depend on hardware performance;
    \item Support for a different number of robots in each team, to enable more environment possibilities;
    \item Split simulated objects collision spaces to create separate collision groups;
    \item Added motor speed constraints matching real-world observations;
    \item Enable cylinder collision, removing the dummy collision object to reduce the total number of simulated bodies;
    \item Defined direct simulator calls in Python for fast communication. Enabling the instantiation of multiple simulators without the need to manage network communication ports.
\end{itemize}

Although the simulator is external to the framework, the simulator module abstracts its interface. Table \ref{tab:simulator_tests} presents the \rsim{} simulator performance in comparison with the grSim simulator in headless mode for a different number of robots on field. The grSim used in the comparison had slight modifications for removing frequency limits, and it also includes a modified version with synchronous operations for comparison. 

\setlength{\tabcolsep}{9pt}
\begin{table}
    \caption{Simulation performance in average and standart deviation of steps per second, for 1, 6 and 11 \gls{ssl} robots in each team.}
    \centering
    \begin{tabular}{lccc}  
        \toprule
        Simulator & 1 vs 1 & 6 vs 6 & 11 vs 11 \\
        \midrule
        grSim (asynchronous) & 2167.9 (8.4) & 408.7 (0.3) & 228.3 (0.1) \\
        grSim (synchronous) & 1894.0 (8.4) & 390.0 (0.5) & 219.0 (0.7) \\
        rSim (proposed, synchronous)& \textbf{2408.8 (9.3)} & \textbf{510.8 (1.8)} & \textbf{288.0 (0.4)} \\
        \bottomrule 
    \end{tabular}
    \label{tab:simulator_tests}
\end{table}

\subsection{Environment Module}
The environment module is where the environment task itself is defined. It implements the interface with the agent and communication with the other framework modules. The interface with the agent complies with the OpenAI Gym \cite{gym} framework, and it communicates with the other modules using the \textit{entities} structures. 

The use of common interfaces enables the definition of base environments, which handle the communications with the other modules and the compliance with Gym. The framework provides benchmark environments of important tasks related to the RoboCup challenges \cite{robocup}, serving as examples and making it easier for other researchers to develop and evaluate new \gls{rl} methods in these benchmark scenarios. The work needed for defining a new environment consists of the implementation of only four methods:

\begin{itemize}
    \item \textbf{get\_commands}: Returns a list of \textit{Robot entities} containing the commands which are sent to the simulator;
    \item \textbf{frame\_to\_observations}: Returns an observation array which will be forwarded to the agent as defined by the environment;
    \item \textbf{calculate\_reward\_and\_done}: Returns both the calculated step reward and a boolean value indicating if the current state is terminal;
    \item \textbf{get\_initial\_positions\_frame}: Returns a \textit{Frame entity} used to define the initial positions of the ball and robots.
\end{itemize}

\subsection{Render Module}
Although we explicitly removed the graphical interface from the simulator for performance, the render module enables visualization without previous drawbacks. It renders on-demand a 2D image of the field and has no performance reduction when not in use. Its implementation is independent of the simulator and enables it to be used at training time for monitoring purposes since it is based on the Gym base environment solution.

\section{Proposed Robot Soccer Environments}
\label{sec:envs}

Due to the differences of the leagues mentioned in Section \ref{sec:intro}, we propose a complete soccer game environment based on Latin American Robotics Competition competition for the \gls{vsss} and simple skills learning environments for the \gls{ssl}.

A \textit{state} is defined as the complete set of data returned by the simulator after a performed action and an \textit{observation} as a subset or transformation of this state. On the following proposed environments, we described the state by positions $(x,y)$, angles ($\theta$), and velocity $(vx,vy,v\theta)$ of each object (ball, teammate, and opponent) in reference to the field center. On the \gls{ssl} environments there is an additional \gls{ir} signal of each robot, indicating if the ball is in contact with the kicking device.

\subsection{IEEE Very Small Size League Environments}
Based on Bassani et al. \cite{bassani2020framework}, we developed a single and a multi-agent benchmark for the \gls{vsss} league. The observation is the complete state defined above. We describe the actions of each robot as the power percentage for each wheel that the robot will apply in the next step. For the non-controlled agents, we use a random policy based on \gls{ou} \cite{OUprocess}. The \gls{ou} process creates a more continuous motion trend for a few steps, which allows the agents to follow a more structured random trajectory instead of just oscillating around the initial point. An episode finishes if the agent received/scored a goal or if the timer reaches 30 seconds of simulation. In the \textbf{IEEE VSSS Single-Agent} environment, only one robot learns a policy, and the other five (two teammates and three opponents) follow a random policy that consists of executing actions sampled according to the \gls{ou} process. In the \textbf{IEEE VSSS Multi-Agent} environment, the controlled robots share the learning policy. See on Fig. \ref{fig:vss_env} the rendered Frame entity of the IEEE \gls{vsss} environments.

\subsection{Small Size League Environments}
\label{sec:sslenvs}
The first environment developed is the basic GoToBall. The other environments were based on RoboCup's 2021 hardware challenge \cite{hw_challenges}.

The actions of the \gls{ssl} environments are the global frame velocities on each axis, kick power, and dribbler state (on/off). For all environments, we defined rewards based on energy spent by the robot, its distance to the ball, and for reaching the objective.

\begin{figure}[!ht]
\begin{center}
    \subfigure[IEEE VSSS]{\includegraphics[width=0.3\textwidth]{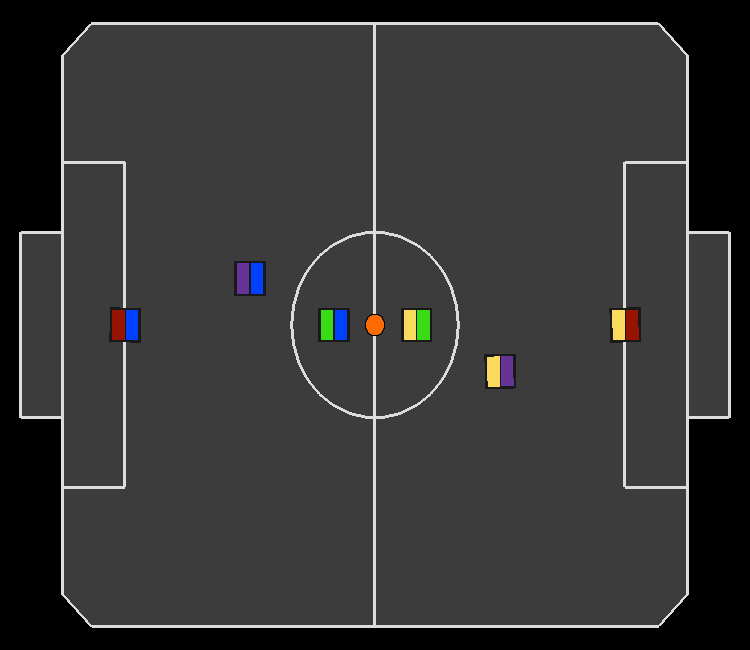} \label{fig:vss_env}}
    \subfigure[GoToBall]{\includegraphics[width=0.3\textwidth]{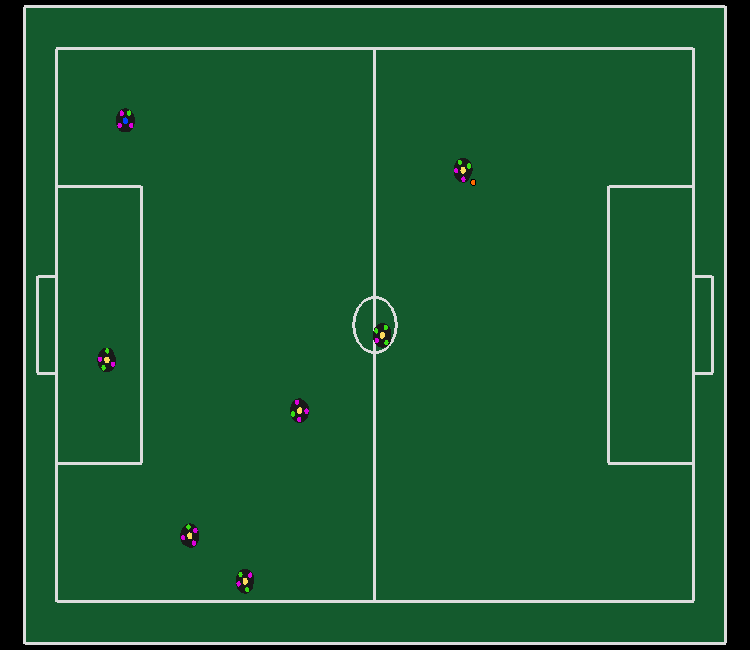} \label{fig:goto_env}}
    \subfigure[Static Defenders]{\includegraphics[width=0.3\textwidth]{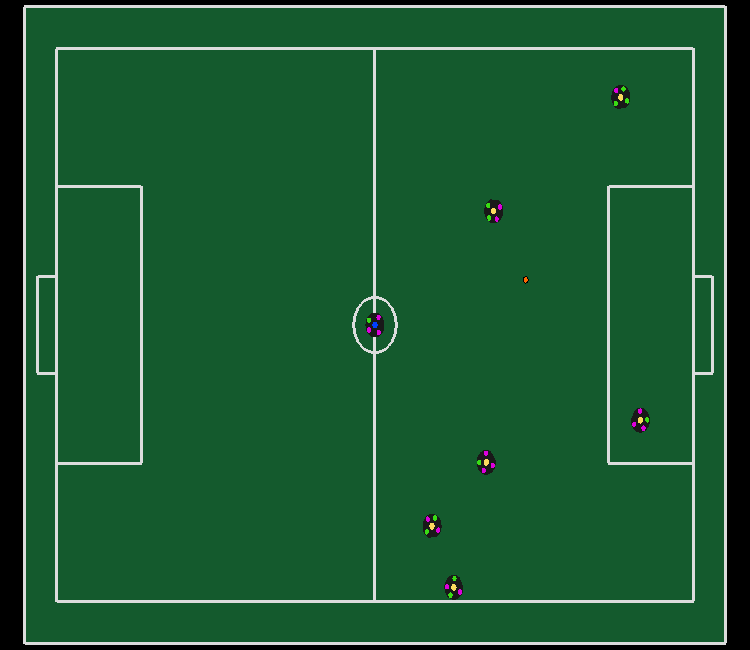} \label{fig:static_env}}
    \subfigure[Contested Possession]{\includegraphics[width=0.3\textwidth]{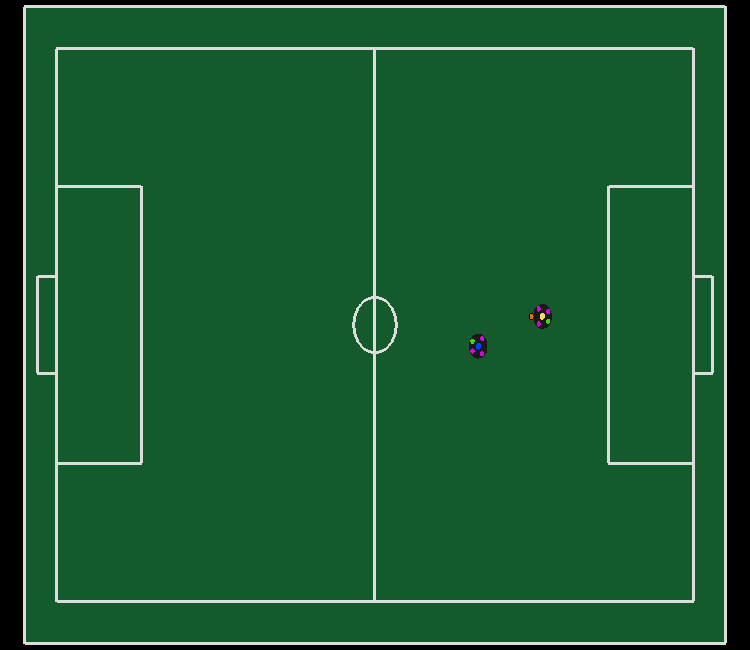} \label{fig:contested_env}}
    \subfigure[Dribbling]{\includegraphics[width=0.3\textwidth]{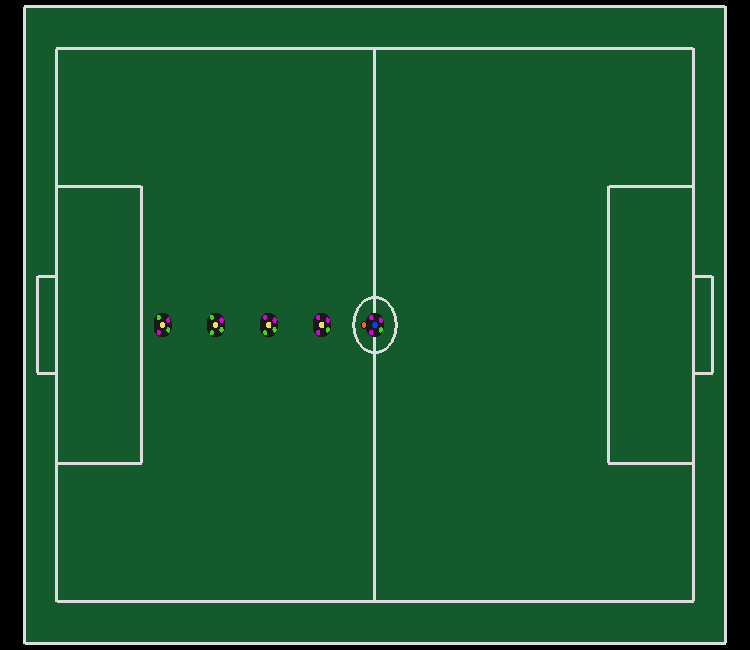} \label{fig:dribbling_env}}
    \subfigure[Pass Endurance]{\includegraphics[width=0.3\textwidth]{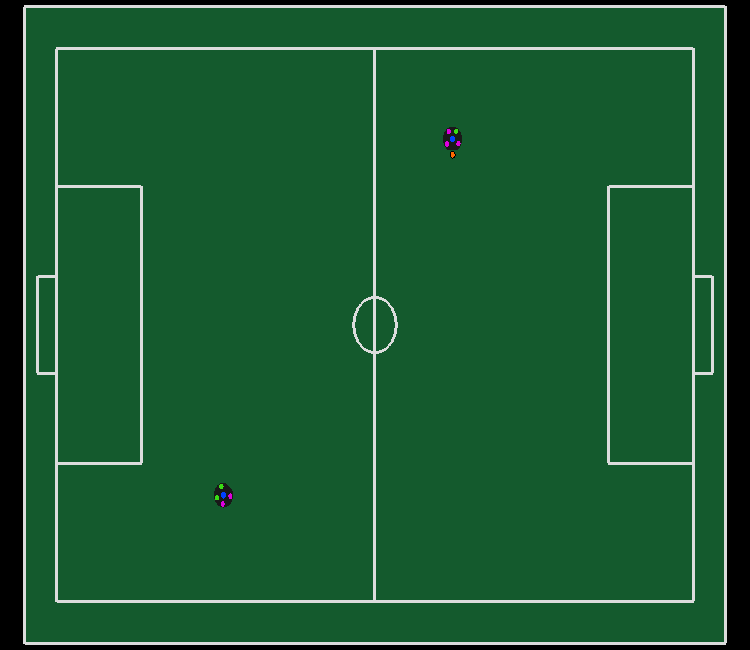} \label{fig:pass_env}}
\end{center}
    \caption{Initial states of the proposed benchmark environments. }
    \label{fig:environments}
\end{figure}

The \textbf{GoToBall} environment is the most straightforward skill to be learned. In this environment, the controlled agent must reach the ball and position its \gls{ir} sensor on it, i.e., arriving at the ball at a certain angle. The episode ends when the robot completes the objective, if the agent exits the field limits, or if the simulation timer reaches 30 seconds. See on Fig. \ref{fig:goto_env} an example of rendered Frame of the environment.

The \textbf{Hardware Challenges} environments consist of four environments based on RoboCup's 2021 hardware challenges. We made certain simplifications to the original environments to make them learnable by the currently available methods in a reasonable amount of time \cite{hw_challenges}. They are:

\begin{enumerate}

    \item \textbf{Static Defenders:} the episode begins with the controlled agent in the field center and 6 opponents and the ball randomly positioned in opponent's field. The episode ends if the agent scores a goal, the ball or the agent exits the opponent's field, the agent collides with an opponent, or the timer reaches 30 simulated seconds. See on Fig. \ref{fig:static_env} an example of initial Frame.

    \item \textbf{Contested Possession:} the episode begins with the controlled agent in the field center and an opponent is randomly positioned in the opponent's field, with the ball on its dribbler. The objective of this challenge is to sneak the ball from the opponent and score a goal. The episode ends with the same conditions of the Static Denfenders environment. See on Fig. \ref{fig:contested_env} an example of initial Frame.

    \item \textbf{Dribbling:} the episode begins with the controlled agent in the field center with the ball on its dribbler and four opponent robots positioned in a sparse row, leaving "gates" between each of them. The objective of this challenge is to dribble the ball while the agent moves through these gates. The episode ends if the agent collides with any robot or exits the field. See on Fig. \ref{fig:dribbling_env} an example of the initial Frame.

    \item \textbf{Pass Endurance (single and multi-agent):} the episode begins with the two robots at random positions, with the ball on the dribbler of one of them. There are no opponents in this environment. In the single-agent environment, the objective is to perform a pass in three seconds. For the multi-agent, they have to perform as many passes as possible in 30 seconds. The episode ends if a pass does not reach the teammate or if the time is out. See on Fig. \ref{fig:pass_env} an example of initial Frame.
\end{enumerate}

\section{Experimental Results} \label{sec:results}
This section presents and discusses the results obtained on our framework with two state-of-the-art deep reinforcement learning methods for continuous control.

\begin{figure}[!ht]
\begin{center}
    \subfigure[IEEE VSSS  Single-Agent]{\includegraphics[width=0.46\textwidth]{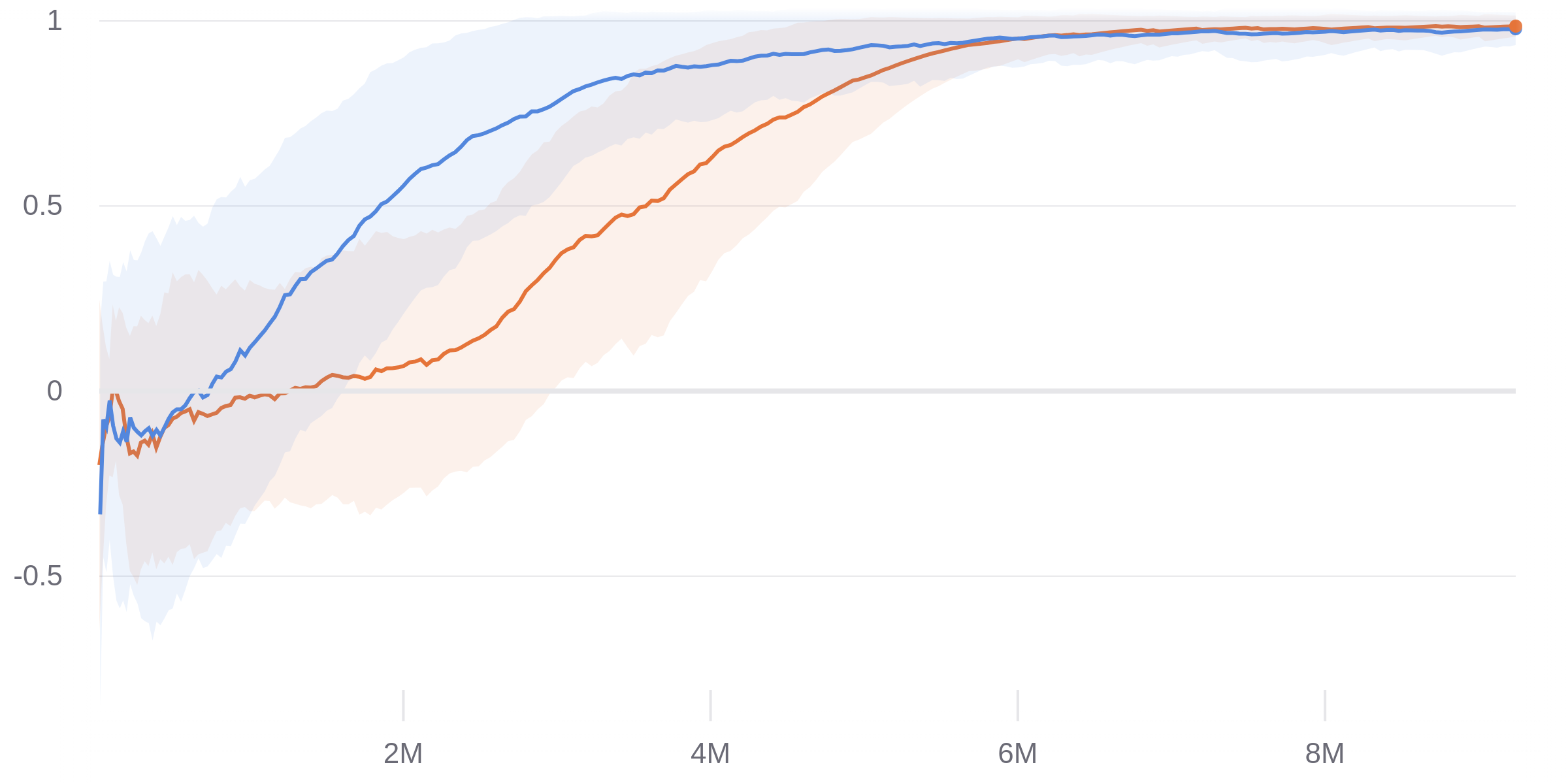} \label{fig:res_a}}
    \subfigure[IEEE VSSS Multi-Agent]{\includegraphics[width=0.46\textwidth]{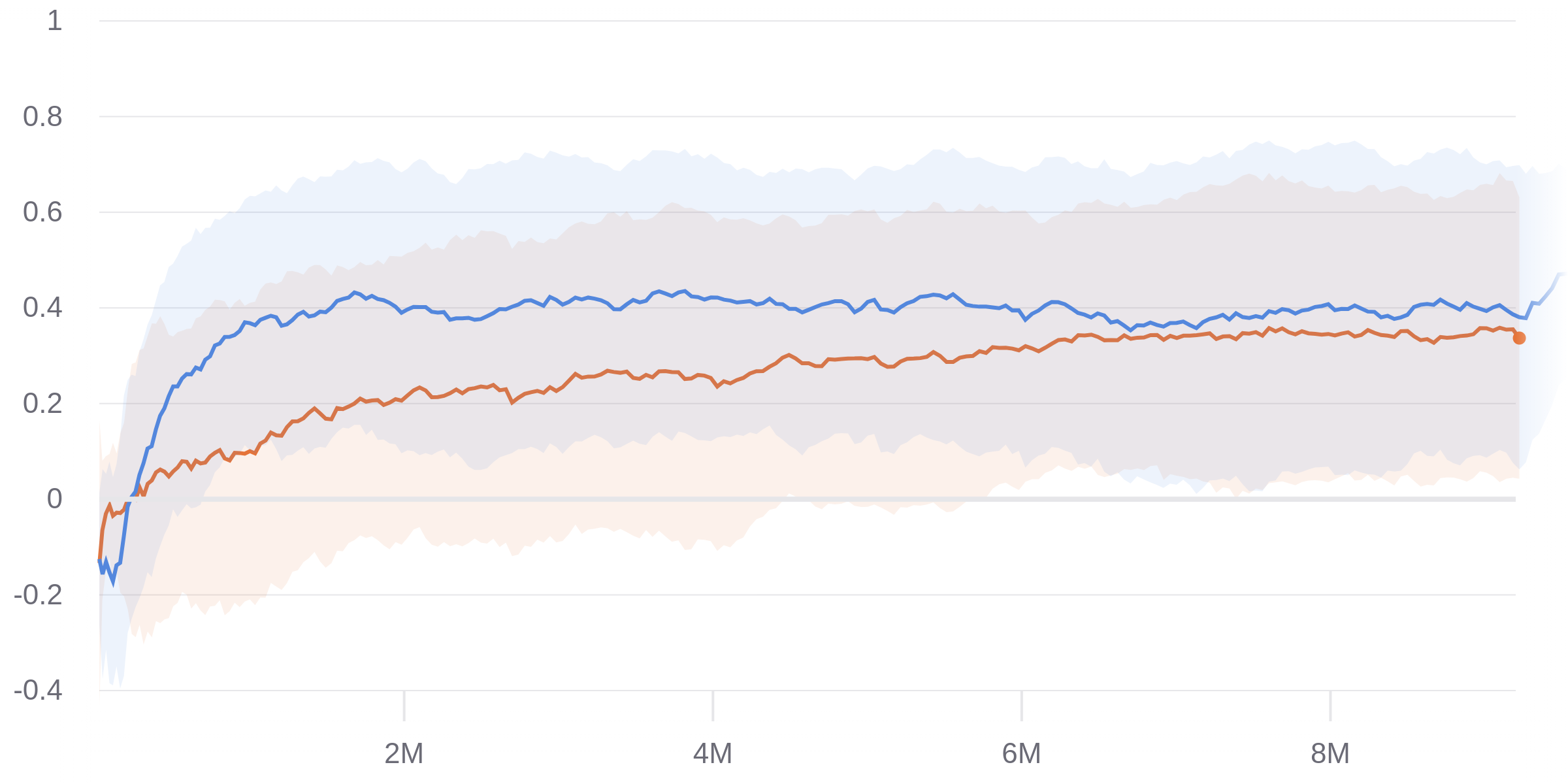}  \label{fig:res_b}}
    \subfigure[GoToBall]{\includegraphics[width=0.46\textwidth]{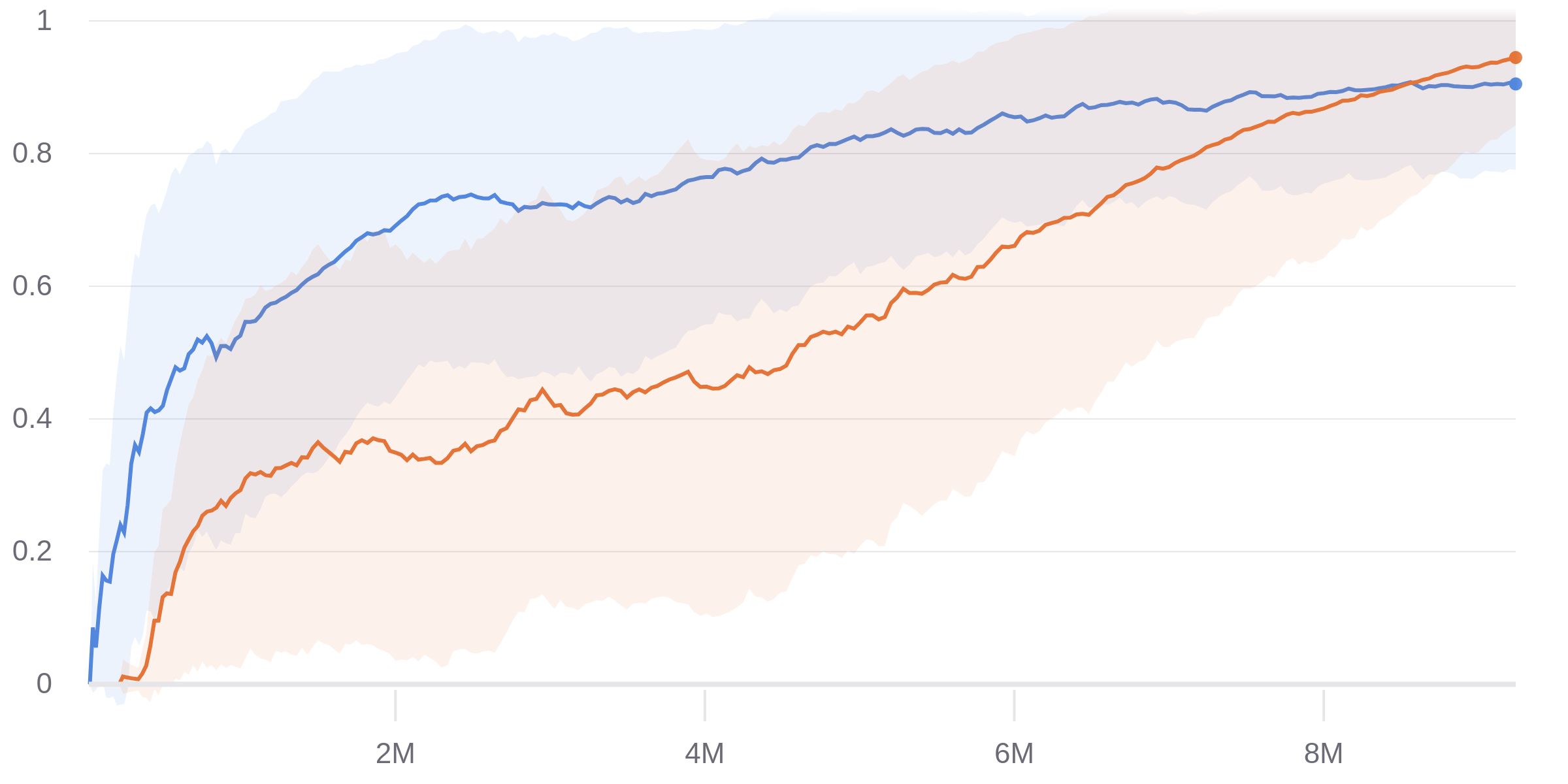}  \label{fig:res_c}}
    \subfigure[Dribbling]{\includegraphics[width=0.46\textwidth]{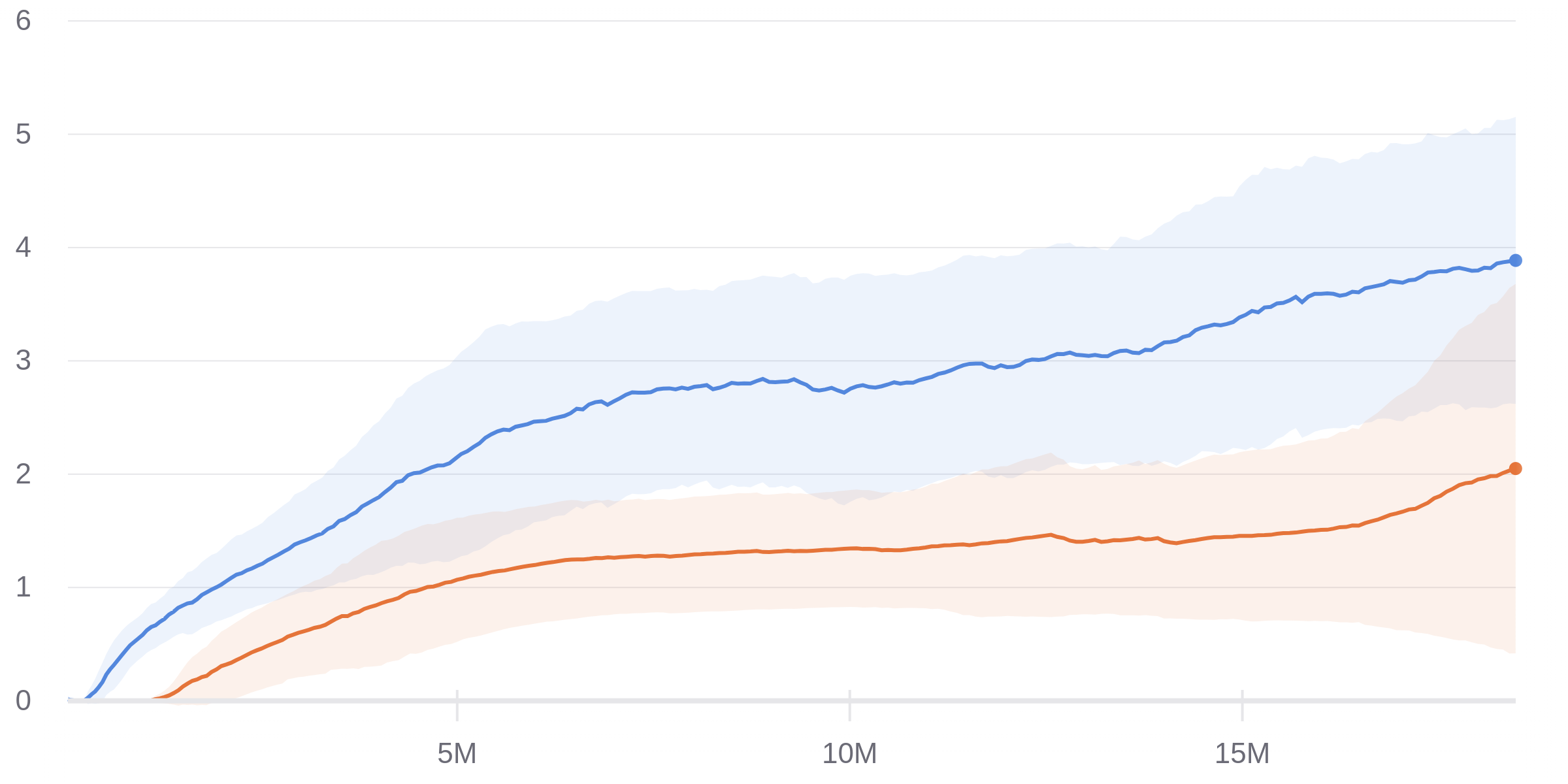}  \label{fig:res_d}}
    \subfigure[Contested Possession]{\includegraphics[width=0.46\textwidth]{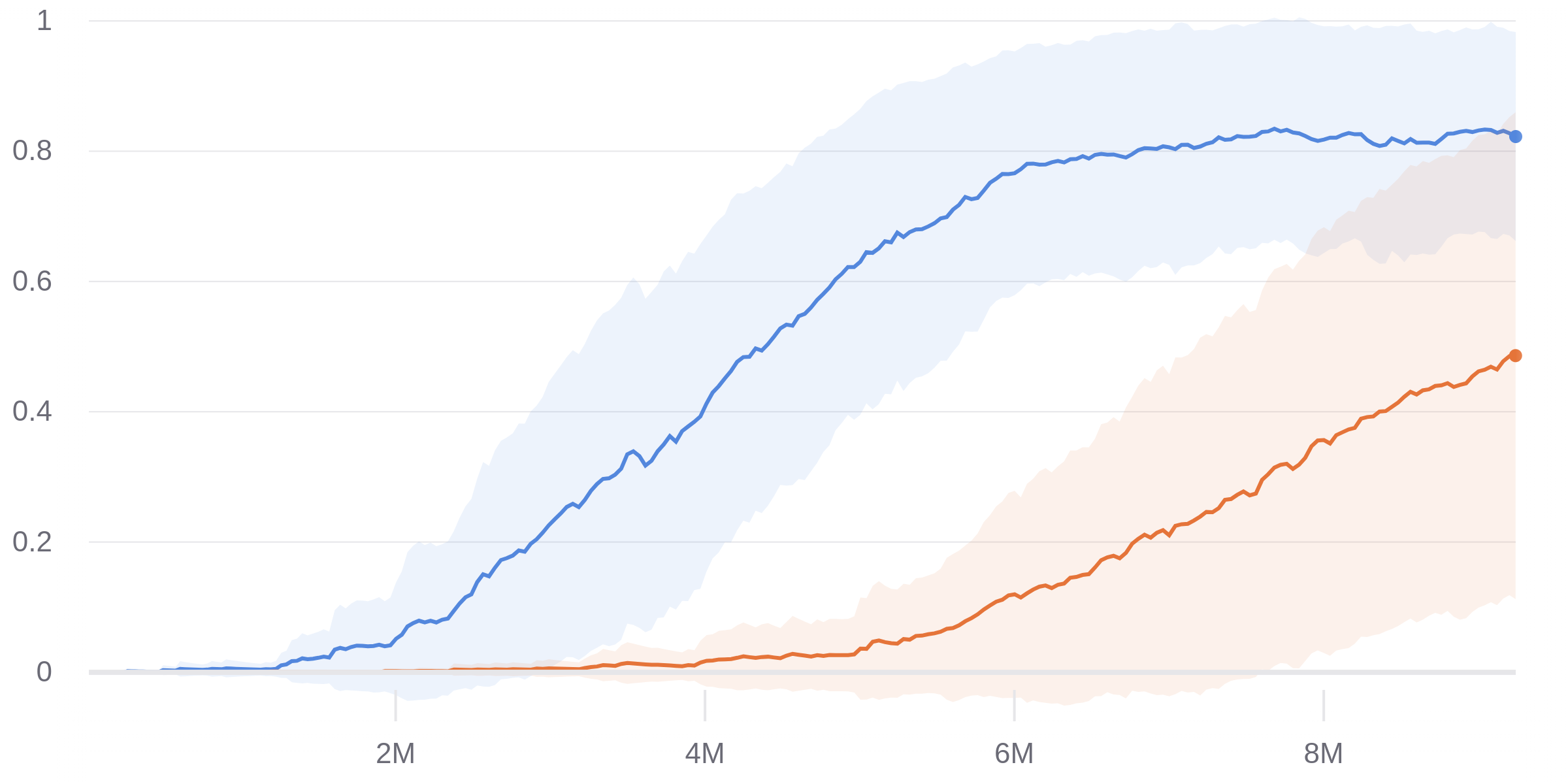}  \label{fig:res_e}}
    \subfigure[Static Defenders]{\includegraphics[width=0.46\textwidth]{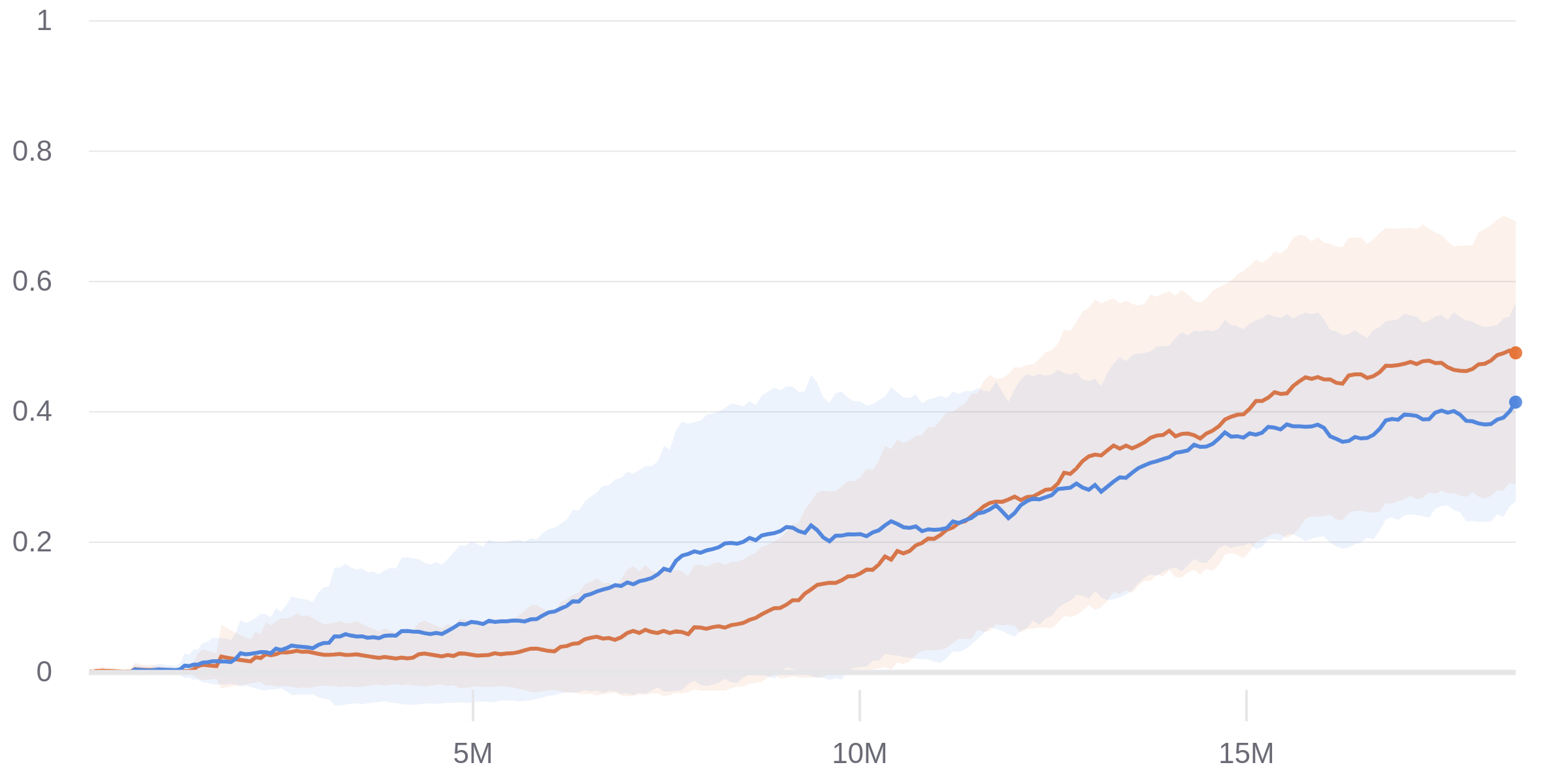}  \label{fig:res_f}}
    \subfigure[Pass Endurance]{\includegraphics[width=0.46\textwidth]{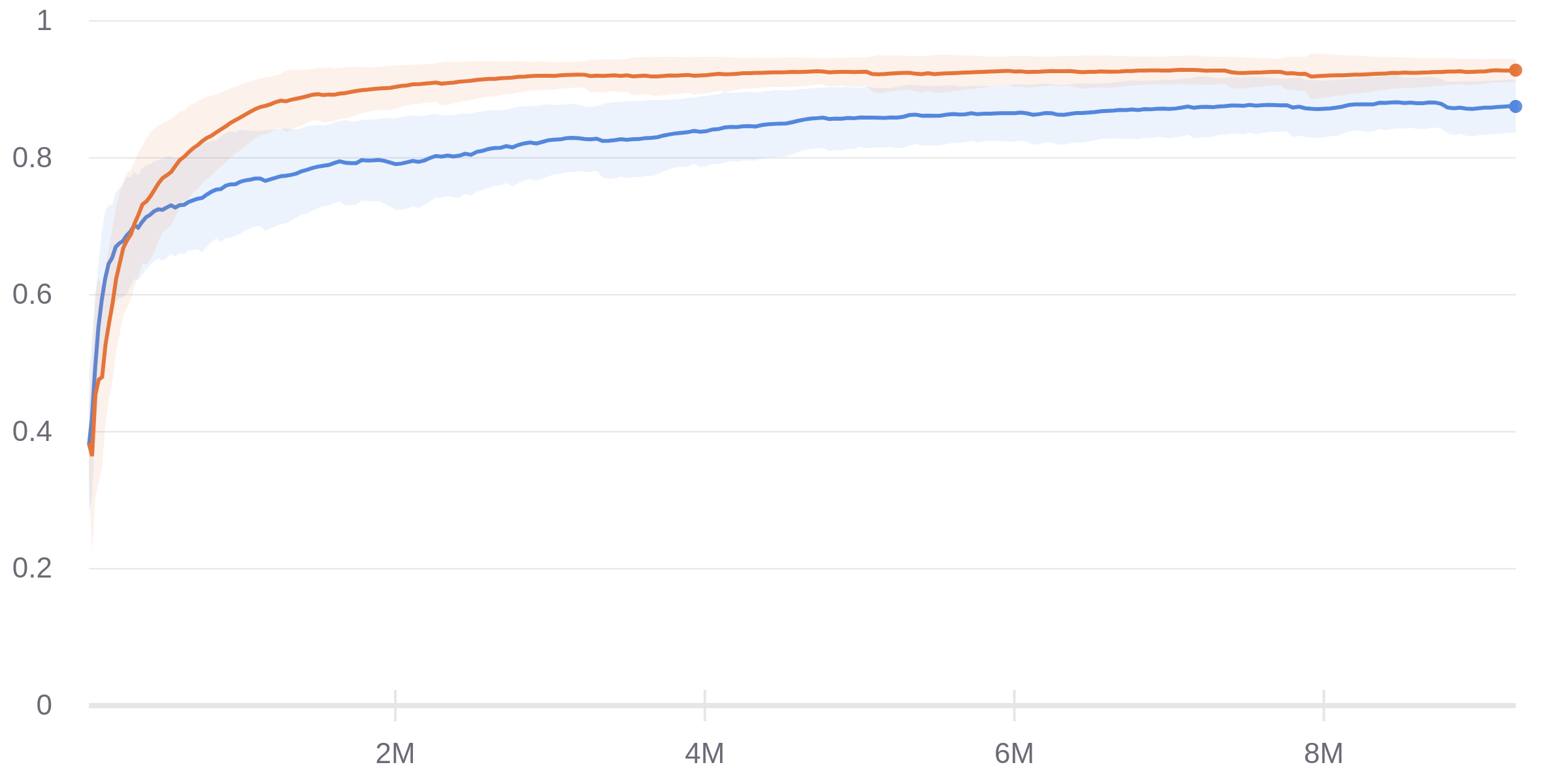}  \label{fig:res_g}}
    \subfigure[Pass Endurance MA]{\includegraphics[width=0.46\textwidth]{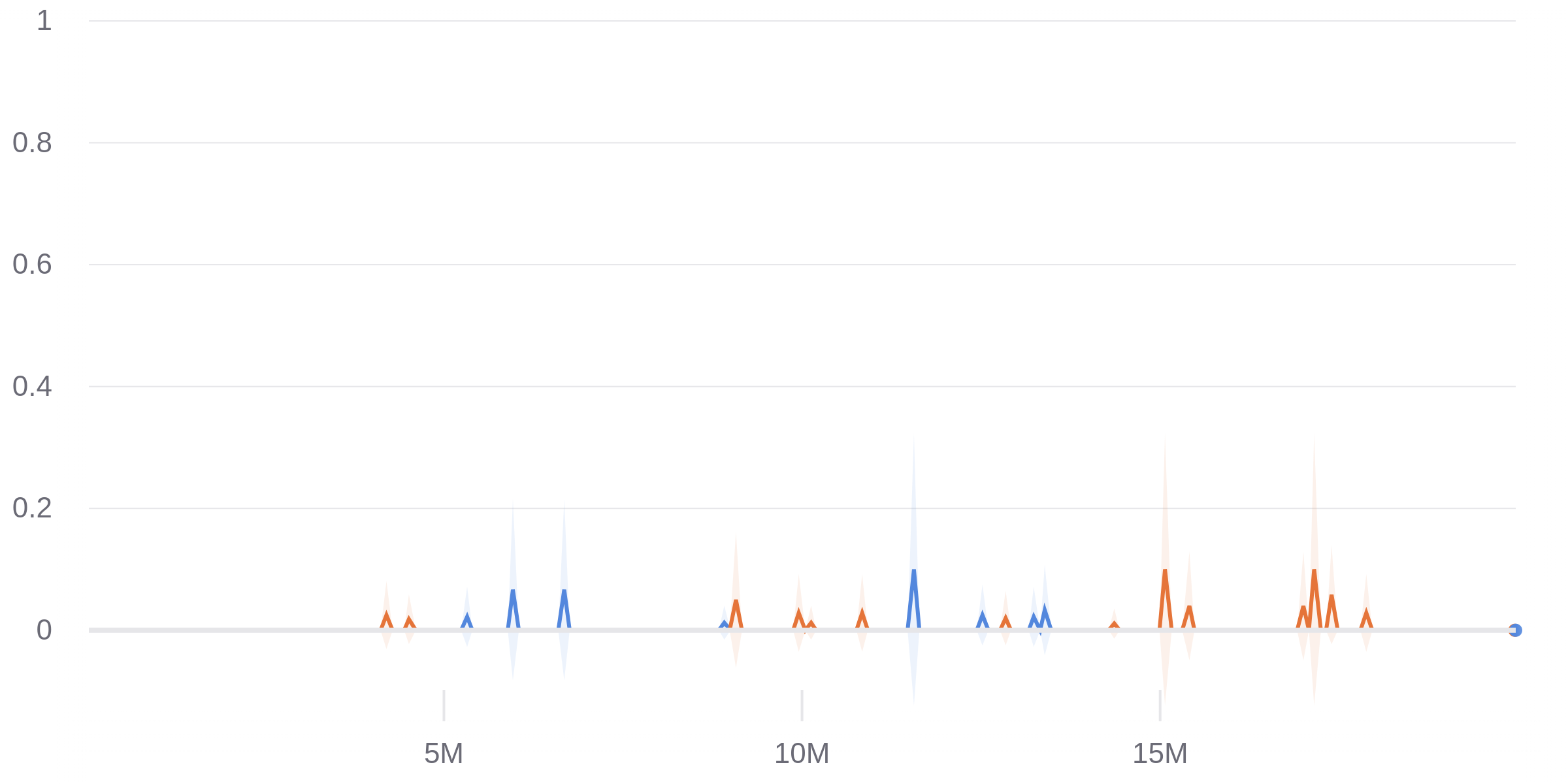}  \label{fig:res_h}}
\end{center}
    \caption{Mean (lines) and standard deviation (shades) of the results obtained for each environment (DDPG in blue and SAC in orange). The Y axis represents: Goal Score for a, b, e, and f; Ball Reached for c; Number of gates transversed for d; Inverse Distance to Receiver for g; and Pass Score for h.}
    \label{fig:results}
\end{figure}

We chose Deep Deterministic Policy Gradient (DDPG) \cite{lillicrap2015continuous} and Soft Actor Critic (SAC) \cite{sac} because both are known for presenting great performance in robot control environments such as Deepmind Control Suite \cite{dmc}. We have also tested Proximal Policy Approximation (PPO) \cite{schulman2017proximal}, however, despite all our efforts in parameter tuning, it was not able to learn even in the easiest environments. Therefore, we concentrated our efforts on DDPG and SAC. On the multi-agent environments, we used a shared policy to control all agents. For each environment, we executed five runs of each method. We ran 10 million steps for each experiment, except for the Dribbling and Static Defenders environments, in which we ran 20 million steps. For the IEEE \gls{vsss} environments, Contested Possession and Static Defenders we use the goal score to evaluate the agents. In the GoToBall, and Dribbling we evaluate if the agents complete or not the respective objective. In the Pass Endurance Single-Agent we used $1/d$ to evaluate the agent, where $d$ is the distance of the ball to the receiver. In the Pass Endurance Multi-Agent we used the pass score to evaluate the agent. In Fig. \ref{fig:results} we present the average and standard deviation of the learning curves obtained with each method in each environment.

We note that both algorithms presented a high standard deviation in all environments, except Pass Endurance. We also point out that DDPG was more sample efficient in most tasks (Figures \ref{fig:res_a} to \ref{fig:res_e}), an interesting result considering SAC usually performs better than DDPG in continuous control environments \cite{sac}. This performance may be explained by the fact that DDPG employs the \gls{ou} process for exploration, which seems to suite better for the environments considered here. As SAC uses an entropy-based exploration, it takes more samples for it to reach the performance of DDPG, although it surpassed DDPG by a small margin at the end, in certain environments (Figures \ref{fig:res_c} and \ref{fig:res_f}).

In the multi-agent environments (IEEE \gls{vsss} and Pass Endurance), we highlight that the results were worse than the respective single-agent ones. This indicates that the agents did not learn to collaborate, since more agents were expected to perform better than a single one. For instance, in the \gls{vsss}, a visual inspection revels that, instead of collaborating, the agents block each other, as can be observed in the frame sequences available in our repository\footnote{\url{https://github.com/robocin/rSoccer}}.

\section{Conclusions and Future Work}
\label{sec:conclusion}

This article presented an open-source framework for developing robot soccer \gls{rl} environments for the \gls{vsss} and \gls{ssl} competitions. The framework includes a simulator optimized for \gls{rl} experiments and an API for defining new environments compatible with the OpenAI Gym standards. It also provides eight benchmark environments that can evaluate \gls{rl} methods regarding different types of robot soccer challenges. The API is easily extensible for other types of environments and tasks. The simulator can be replaced by an interface with real robots for evaluating Sim-to-Real as in \cite{bassani2020framework}.

With this, we aim to put forward research and application \gls{rl} methods for robot soccer by making it easier for other researchers to evaluate their strategies and compare the results in standardized scenarios, therefore improving reproducibility.

Although our results are promising in certain tasks, achieving better results than we would be able to achieve with traditional handcrafted methods, it also makes it clear that much research is needed to achieve an effective robot soccer team trained end-to-end by reinforcement learning. Studying why PPO performed so poorly is essential for our future works, once it achieved interesting results on other studies. The multi-agent (Figures \ref{fig:res_b}, \ref{fig:res_f}) and the Static Defenders Fig. \ref{fig:res_h} environments show that certain benchmarks are too difficult for the currently available methods, indicating an open area of research. In the Static Defenders environment, the best reward function we developed seems inadequate when the dimensionality of the observations increases, hence the poor results. In multi-agent environments, the methods could not learn to cooperate using a shared policy. However, we believe that multi-agent specific algorithms focusing on collaboration such as MADDPG \cite{lowe2017multi} would improve the results.


\section*{ACKNOWLEDGMENTS}

The authors would like to thank RoboCIn - UFPE Team and Mila - Quebec Artificial Intelligence Institute for the collaboration and resources provided; Conselho Nacional de Desenvolvimento Cientifico e Tecnológico (CNPq), and Coordenação de Aperfeiçoamento de Pessoal de Nível Superior (CAPES) for financial support. Moreover, the authors also gratefully acknowledge the support of NVIDIA Corporation with the donation of the RTX 2080 Ti GPU used for this research.
\bibliographystyle{splncs04}
\bibliography{default_content/bibliography}

\end{document}